\documentclass[10pt,3p,authoryear]{elsarticle}
\usepackage{caption}
\usepackage{subcaption}
\usepackage{multirow}
\usepackage{amsmath} 
\usepackage{microtype}
\usepackage{wrapfig}

\journal{Expert Systems With Applications}

\begin{document}

\begin{frontmatter}

\title{Class Distance Weighted Cross Entropy Loss for Classification of Disease Severity}

\author[inst1]{Gorkem Polat}
\ead{gorkem.polat@metu.edu.tr}
\author[inst1]{\corref{cor1}Ümit Mert Çağlar}
\ead{mecaglar@metu.edu.tr}
\cortext[cor1]{Corresponding Author}
\author[inst1]{Alptekin Temizel}
\ead{atemizel@metu.edu.tr}
\affiliation[inst1]{organization={Graduate School of Informatics}, 
addressline={Middle East Technical University},
city={Ankara},
country={Türkiye}}

\begin{abstract}
Assessing disease severity with ordinal classes, where each class reflects increasing severity levels, benefits from loss functions designed for this ordinal structure. Traditional categorical loss functions, like Cross-Entropy (CE), often perform suboptimally in these scenarios. To address this, we propose a novel loss function, Class Distance Weighted Cross-Entropy (CDW-CE), which penalizes misclassifications more severely when the predicted and actual classes are farther apart. We evaluated CDW-CE using various deep architectures, comparing its performance against several categorical and ordinal loss functions. To assess the quality of latent representations, we used t-distributed stochastic neighbor embedding (t-SNE) and uniform manifold approximation and projection (UMAP) visualizations, quantified the clustering quality using the Silhouette Score, and compared Class Activation Maps (CAM) generated by models trained with CDW-CE and CE loss. Feedback from domain experts was incorporated to evaluate how well model attention aligns with expert opinion. Our results show that CDW-CE consistently improves performance in ordinal image classification tasks. It achieves higher Silhouette Scores, indicating better class discrimination capability, and its CAM visualizations show a stronger focus on clinically significant regions, as validated by domain experts. Receiver operator characteristics (ROC) curves and the area under the curve (AUC) scores highlight that CDW-CE outperforms other loss functions, including prominent ordinal loss functions from the literature.

\end{abstract}


\end{frontmatter}

\section{Introduction}
%\linenumbers
In ordinal classification tasks, it is common to use the distance relation between different classes. This relation is evident in many examples, including disease severity classification tasks where classes can correspond to the severity of disease, and their ordered nature can be exploited with custom loss functions \citep{cao2020rank,shi2023deep,belharbi2019unimoconstraints,albuquerque2021ordinal}, instead of employing categorical loss functions.

In the medical domain, disease classes are assigned according to the symptom severity, thus implying an ordinal relation between classes \citep{KELLEY2014119}. Misclassification between further separated classes is more detrimental than the misclassification of relatively closer classes. For example, in the context of a medical diagnosis tool, erroneously categorizing a patient as healthy when they have a severe condition is considerably more problematic than misclassification between a mild or moderate condition. On the other hand, commonly used loss functions consider each class in isolation and do not account for any distance information between them, and both misclassification cases will have the same impact on the error. Consequently, both types of misclassifications are accorded equal weight in terms of error, disregarding the practical impact. To address this, a possible approach is mathematically representing the distance information in a loss function, thereby penalizing misclassifications between classes that are further apart more severely than those closer.

In our earlier work, we investigated application of ordinal loss functions for ulcerative colitis severity estimation \citep{Gorkem_CDWCE}. In this work, we extend the proposed loss function, Class Distance Weighted Cross Entropy (CDW-CE) and demonstrate its capabilities. In addition, we expand the CDW-CE by introducing an additive margin term into the loss function, analyzing class discrimination properties of models trained with different loss functions, and demonstrating its effectiveness with supplementary experiments comparing against various loss functions in the literature.

Ulcerative Colitis is a disease for which several deep learning-based models have been proposed to classify disease severity. In this problem, disease severity is represented in increasing sequential order, and classification models are trained to predict the disease severity classes corresponding to input images correctly. Labeled Images for Ulcerative Colitis (LIMUC) \citep{LIMUC} dataset is a publicly available dataset with ordinal classes and imbalanced class distributions. We have used this dataset to comparatively evaluate CDW-CE and other standard loss functions namely Cross-Entropy (CE), Mean Squared Error (MSE), CORN framework \citep{shi2023deep}, cross-entropy with an ordinal loss term (CO2) \citep{albuquerque2021ordinal}, and ordinal entropy loss (HO2) \citep{albuquerque2021ordinal} from the literature.

We used t-distributed stochastic neighbor embedding (t-SNE) and uniform manifold approximation and projection (UMAP) plots to analyze the latent representations of models trained with CE, CO2, HO2 and CDW-CE. To evaluate the quality of these representations, we calculated the silhouette score, a metric that quantifies inter-class compactness and intra-class separation \citep{MALEPATHIRANA2022104749}, based on the t-SNE and UMAP plots. Additionally, Class Activation Map (CAM) outputs from CDW-CE and CE methods, alongside with raw endoscopic images, were presented to medical experts to gather their feedback on the alignment of CAM outputs with observed symptoms. We also employed receiver operating characteristics (ROC) curves and area under curve (AUC) performance measures to highlight the importance of the loss functions on model performance.

The main contribution of this paper is the introduction of an advanced ordinal loss function, Class Distance Weighted Cross-Entropy (CDW-CE). The proposed CDW-CE loss function outperforms other loss functions in classification performance, feature extraction and explainability with default power and margin values.

\section{Related Work}

Deep learning-based disease severity classification is a widely researched topic and various methods have been proposed. \citet{ALYAMANI20241129} employs conventional convolutional neural networks (CNN) with categorical loss functions for the Ulcerative Colitis classification task. \citet{BARBEROGOMEZ2025128878} emphasizes the limited explainability of CNN methods and employs class activation maps to improved interpretability. \citet{DELATORRE2018144} explores the use case of quadratic weighted kappa (QWK) as a loss function for ordinal data.

Ordinal categories are frequently observed in real-world situations \citep{unal2021new}, particularly in the medical domain \citep{seveso2020ordinal}, such as disease severity classifications \citep{BARASH2021187}. For ordinal classification problems, tailored methodologies, including novel loss functions that take ordinality into account, are reported to outperform traditional methods \citep{TANG2023126245}. 
\citet{BENDAVID2008825} emphasizes the importance of employing performance metrics, such as Cohen's Kappa, for ordinal and multi-class classification problems. \citet{KAYADIBI2023120617} propose using ROC curves and corresponding AUC scores to evaluate deep learning models trained on imbalanced datasets, which are common in the medical domain. \citet{BATCHULUUN2023119809} emphasize the use of class activation maps (CAMs) in deep learning models, particularly convolutional neural networks, to identify feature-level activations associated with the trained model's attention. Similarly, \citet{AHAMED2024124908} employ CAM visualizations for better explainability, however, they do not employ medical domain expert feedback, leaving it for future studies.

This section provides an overview of methods that utilize such loss functions and evaluation methods.

An end-to-end deep learning method for ordinal regression, focusing on age estimation from facial images, was introduced by \citet{niu2016ordinal}. This approach involved transforming ordinal regression into a series of binary classification sub-problems and employing a multiple-output CNN for joint task resolution. By using $N-1$ output nodes for $N$ classes and applying label extension to the target class, their method is reported to perform better than other ordinal regression techniques, such as metric learning. However, this approach exhibited inconsistencies in output ranking within the subtasks. Consistent Rank Logits (CORAL) framework \citep{cao2020rank} addresses rank-inconsistencies in extended binary classification models. This framework tackles the problem of inconsistent classifier predictions in neural networks, where model confidence should ideally decrease as the rank increases. CORAL differs from the method proposed by \citet{niu2016ordinal}, employing weight sharing in the penultimate layer during training, ensuring classifier consistency. The authors demonstrated the effectiveness of CORAL when integrated into common CNN architectures like ResNet, resulting in enhanced predictive performance for age estimation tasks. Conditional Ordinal Regression for Neural Networks (CORN) framework \citep{shi2023deep} relaxes the constraints on the penultimate layer of the CORAL framework and incorporates conditional probabilities. Experiments on various datasets, including MORPH-2, AFAD, AES, and FIREMAN, showed that CORN outperformed previous methods. However, as mentioned earlier, these techniques need modification of both model architecture and labeling structure. An alternative approach for ordinal classification is integrating uni-modality distribution into the model's output predictions, penalizing inconsistencies in posterior probability distribution between neighboring labels alongside the primary loss function, often using cross-entropy.

Non-parametric ordinal loss \citep{belharbi2019unimoconstraints} encourages output probabilities to follow an unimodal distribution by penalizing inconsistent probability pairs in the output. This approach was validated on various tasks, including breast cancer grading, predicting the decade of a photograph, and age estimation. An extension of this technique by adding unimodality losses to cross-entropy (CO2) and entropy losses (HO2) for the final loss function was proposed by \citet{albuquerque2021ordinal}, and it was validated using the Herlev dataset \citep{jantzen2006analysis}, consisting of cervical cell images, and various CNN architectures. Both studies reported improved performance compared to the method by \citet{niu2016ordinal} and cross-entropy loss.

An alternative approach involves using regression to estimate a single continuous value at the output or applying a sigmoid activation function to constrain predictions within the range of [0, 1], which are then transformed into discrete levels using thresholds or probability distributions. \citet{beckham2016simple} proposed adding a single-node layer on top of the final layer with squared-error loss, suggesting that processing the final output through a sigmoid function followed by multiplication by (K - 1) yields better results, with rounding to the nearest integer for inference. They reported improved performance compared to standard cross-entropy loss. \citet{10.1093/ibd/izac226} developed a regression-based deep learning system, leveraging convolutional neural networks for enhanced accuracy and robustness in classifying Ulcerative Colitis severity. However, It is worth noting that regression-based approaches have generally been reported to exhibit inferior performance in various studies \citep{albuquerque2021ordinal, belharbi2019unimoconstraints}.

\section{Methods}
This work introduces an ordinal loss function designed to improve model performance in ordinal classification tasks, as previously discussed. This section provides detailed information on developing our custom loss function, the experimental setup, evaluation metrics and the dataset used.

\begin{figure}[ht!]
  \centering
  \includegraphics[width=0.6\textwidth,keepaspectratio]{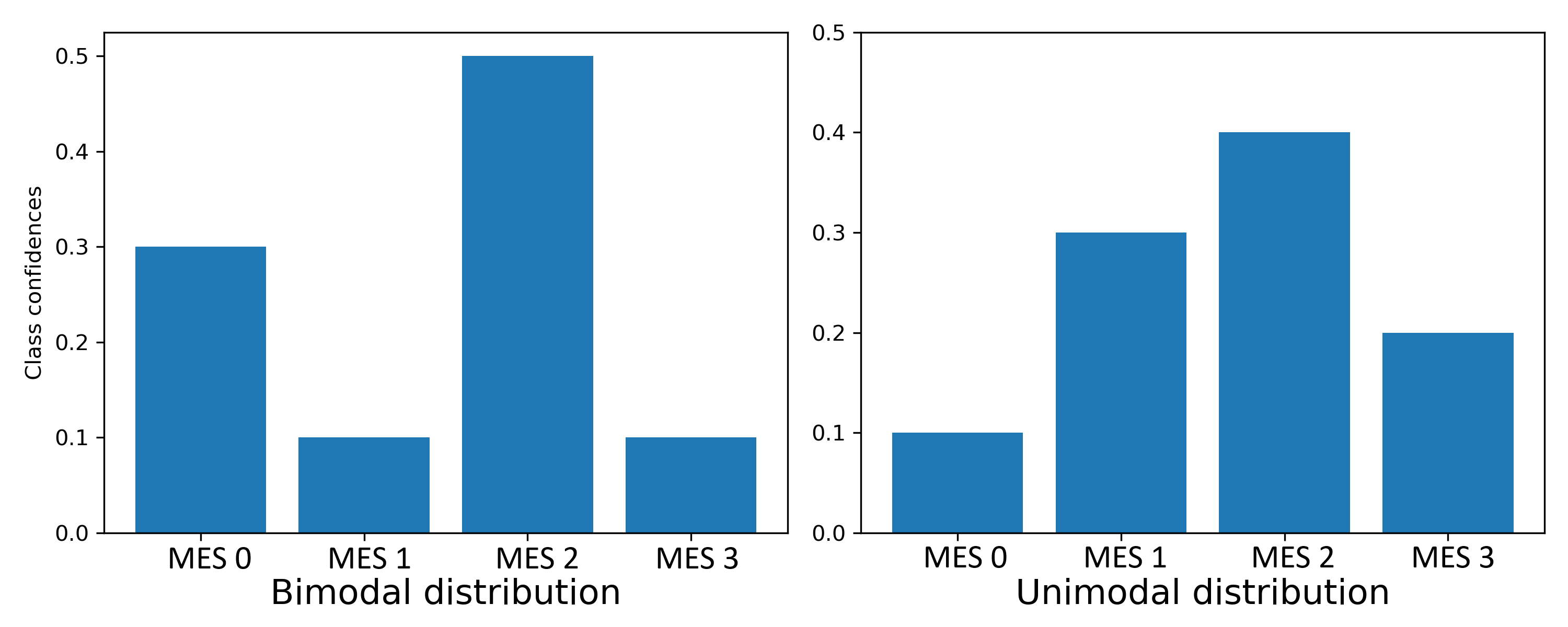}
  \caption[CE loss function and unimodal distribution.]{Assuming MES-2 is the true class, the loss calculated with CE is the same for both cases. On the other hand, the unimodal distribution is more intuitive for ordinal classification.}
  \label{fig:unimodal_vs_bimodal}
\end{figure}

The main idea behind employing an ordinal loss function is demonstrated in Figure \ref{fig:unimodal_vs_bimodal}. Unlike categorical loss functions, such as cross-entropy loss, which treat bimodal and unimodal distributions the same, an ordinal loss function can address issues specific to ordinal datasets. Mayo Endoscopic Scoring (MES) is used to classify the severity of Ulcerative Colitis, an inflammatory bowel disease. MES classes are ordinal, and thus, the misclassification of MES-3 (severe disease conditions) as MES-0 (no disease, healthy) is comparatively more harmful than the misclassification of adjacent classes (moderate and mild disease severities). We aim to encourage deep learning models to predict classes correctly while resembling an unimodal distribution to improve the robustness of the predictions.

\subsection{Class Distance Weighted Cross Entropy Loss}

We start with the standard cross-entropy loss (Eq. \ref{eqn:CE}) as the baseline of our method.

\begin{equation} \label{eqn:CE}
    \textrm{CE}=-\sum_{i=0}^{N-1}{y_i \times \log{\hat{y}_i}}=-\log{\hat{y}_{c}}
\end{equation}

\noindent In Equation \ref{eqn:CE}, $i$ refers to the index of the class in the output layer, $c$ is the index of the ground-truth class, $y$ is the ground-truth label, and $\hat{y}$ refers to the prediction. Since one-hot encoding is used for ground-truth labels, $y_i$ becomes $0$  for $\forall i\neq c$; therefore, CE loss only calculates the negative log-likelihood of the confidence of the ground-truth class (This is why CE is sometimes called the negative log-likelihood function).

The proposed Class Distance Weighted Cross Entropy (CDW-CE) (Eq. \ref{eqn:CDW-CE}) is a non-parametric loss function for multi-class classification. This function penalizes misclassifications more severely when they diverge farther from the true class. The addition of a class distance penalty term aims to regulate the penalty of class misprediction based on the distance between the classes.

Moreover, including a margin term encourages tighter grouping within the same class while enhancing separation between different classes. CDW-CE comprises two main parameters: a logarithmic loss term indicating the extent of deviation for each misclassification from the correct value and a coefficient term that considers the distance to the ground-truth class. The parameter $\alpha$ in the loss coefficient governs the intensity of the penalty, where higher values lead to harsher penalties for misclassifications that significantly deviate from the true class.

\begin{equation} \label{eqn:CDW-CE}
\centering
    \textrm{CDW-CE} = -\sum_{i=0}^{N-1} {\log(1-\hat{y}_i) \times |i-c|^{\alpha}}
\end{equation}

CDW-CE loss is differentiable and can be used directly in backpropagation calculations. As shown in Equation \ref{eqn:CDW-CE-derivative}, the derivative term increases when the prediction value deviates further from its target value of zero.

\begin{equation} 
\label{eqn:CDW-CE-derivative}
\frac{d(L)}{d(\hat{y}_i)} = \frac{|i-c|^{\alpha}}{1-\hat{y}_i}
\end{equation}

Adding a margin to the loss function enforces intra-class compactness and increases inter-class distances, as shown in ArcFace \citep{deng2019arcface}, CosFace \citep{wang2018cosface}, SphereFace \citep{liu2017sphereface}, and NormFace \citep{wang2017normface}. This is incorporated into CDW-CE loss by adding a margin to the class probabilities. CDW-CE with margin loss (Equation \ref{eqn:CDW-CE-with-margin}) is expected to improve the overall classification performance. 

\begin{equation} 
\label{eqn:CDW-CE-with-margin}
    \textrm{\textbf{CDW-CE with Margin}} = -\sum_{i=0}^{N-1} {\log(1- max(1, \hat{y}_i + m)) \times |i-c|^{\alpha}}
\end{equation}

CO2 and HO2 losses, shown in Eqs. {\ref{eqn:CO2}} and {\ref{eqn:HO2}} respectively, are frequently used for comparison in this study. In these equations, CE and H refer to Cross entropy and Entropy losses, respectively, $y_n$ is ground truth, $\hat{y}_n$ is model predictions, $K$ is the total number of classes, $\lambda$ is a parameter that determines the strength of the unimodal loss, and $\delta$ is the margin term.

\begin{align} 
\label{eqn:CO2}
\begin{split}
    \text{CO2} \left(y_n,{\hat{y}}_n\right) = \text{CE} \left(y_n,{\hat{y}}_n\right) + \lambda\sum_{k=0}^{K-1}{\mathbf{1}(k\geq k_n^\ast)\text{RELU}(\delta+{\hat{y}}_{n\left(k+1\right)}-{\hat{y}}_{n(k)})}+ \\ 
    \lambda\sum_{k=0}^{K-1}{\mathbf{1}(k\le k_n^\ast)\text{RELU}(\delta+{\hat{y}}_{n\left(k\right)}-{\hat{y}}_{n(k+1)})}
\end{split}
\end{align}

\begin{align} 
\label{eqn:HO2}
\begin{split}
    \text{HO2} \left(y_n,{\hat{y}}_n\right) = \text{H} \left(y_n,{\hat{y}}_n\right) + \lambda\sum_{k=0}^{K-1}{\mathbf{1}(k\geq k_n^\ast)\text{RELU}(\delta+{\hat{y}}_{n\left(k+1\right)}-{\hat{y}}_{n(k)})}+ \\ 
    \lambda\sum_{k=0}^{K-1}{\mathbf{1}(k\le k_n^\ast)\text{RELU}(\delta+{\hat{y}}_{n\left(k\right)}-{\hat{y}}_{n(k+1)})}
\end{split}
\end{align}

By employing a power term for the distance of misclassification, the CDW-CE encourages the model to correctly form a unimodal prediction similar to CO2 and HO2. However, the power term penalizes further misclassifications more harshly thus preventing the model to favor majority-classes.

\subsection{Evaluation Metrics}

Evaluating the performance of deep learning methods is a challenging task that requires different approaches and solutions \citep{YILMAZ2023110020}. The classification performance of deep learning models was evaluated using accuracy, F1 score and quadratic weighted kappa (QWK). Additionally, ROC curves and their corresponding area under the curve(AUC) scores as well as confusion matrices were analyzed to compare different loss functions. Evaluation metrics used in this work are summarized with their associated descriptions and equations as well as widely used abbreviations in Table \ref{tab:evals}.

%%NEW

\begin{table}
\centering

    \begin{tabular}{|p{0.15\linewidth} | p{0.2\linewidth} | p{0.55\linewidth}|}\hline
\textbf{Metric}      & \textbf{Abbreviation} & \textbf{Explanation}                         \\\hline\hline    
True \newline Positive  & TP & Correct prediction of a positive class                        \\\hline 
True \newline Negative  & TN & Correct prediction of a negative class                        \\\hline 
False \newline Positive & FP & Incorrect prediction of a positive class \newline (type-1 error)       \\\hline 
False \newline Negative & FN & Incorrect prediction of a negative class \newline(type-2 error)       \\\hline 
Accuracy       &    & Ratio of correct predictions to all samples \newline (TP+TN/TP+TN+FP+FN) \\\hline 
Precision      &    & Ratio of correct predicted positive samples to all positive predictions (TP/TP+FP)\\\hline 
Recall         &    & Ratio of correct predicted positive samples to actual positive samples (TP/TP+FN)\\\hline 
F1 Score       &    & Harmonic mean of precision and recall \newline (2$*$(Recall$*$Precision)$/$(Recall$+$Precision)\\\hline 
Kappa          &    & Ratio of observed accuracy to expected accuracy of random probability \\\hline 
Quadratic \newline Weighted \newline Kappa  &  QWK  & Kappa metric with more weight to further misclassifications between predicted and actual samples\\\hline 
Mean\newline Absolute\newline Error       & MAE  & The average absolute difference between predicted and actual samples\\\hline 
    \end{tabular}
    \caption{Evaluation metrics used.}
    \label{tab:evals}
    
\end{table}

%%NEW

In addition to the classification performance metrics, we have also calculated remission scores, a score used by medical experts that bins lower MES classes of 0 and 1 and higher MES classes of 2 and 3 together. This remission score is essential for domain experts aiming to classify whether a case is healthy or contains mild symptoms, which corresponds to the remission class, or contains moderate to severe symptoms, which is the non-remission class. We report the comparative performance evaluation of various methods using Quadratic Weighted Kappa (QWK) for multi-class classification, Kappa for binary (remission) classification, and accuracy, and F1 scores.

The classification task followed a one-vs-rest approach where the class with the highest probability was selected as the prediction. The deep learning models were also adapted for regression by modifying their final layers to have a single output node, without an activation function, and using the MSE loss function. This approach allows for the evaluation of regression-based models in this work. Mean absolute error (MAE), a typical baseline performance metric for regression tasks, was reported to compare regression based classification methods.

\subsection{Experimental Setup}
%%NEW
\begin{figure}[htbp]
  \centering
  \includegraphics[width=0.9\textwidth]{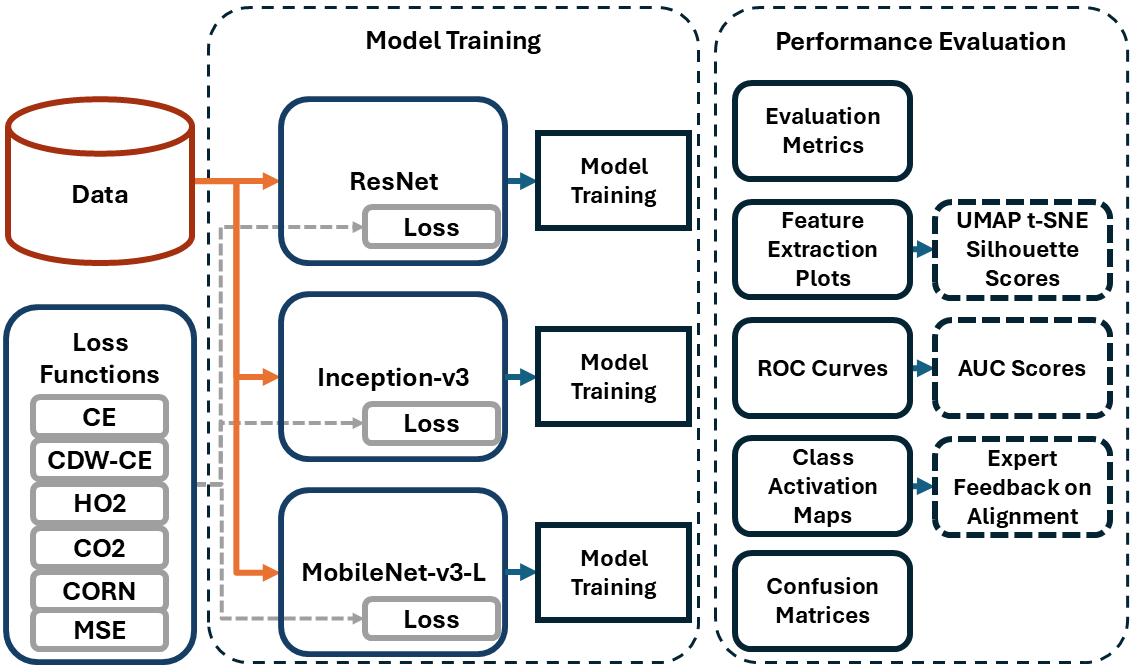}
  \caption{The flowchart of the experimental setup.}
  \label{fig:flowchart}
\end{figure}
%%NEW

We used LIMUC \citep{LIMUC} dataset and employed ResNet18 \citep{he2016deep}, Inception-v3 \citep{Szegedy_2016_CVPR}, and MobileNet-v3-large {\citep{howard2019searching}} architectures for experimental evaluation. We have implemented the proposed ordinal loss function Class Distance Weighted Cross Entropy (CDW-CE) and other loss functions from the literature such as Cross-Entropy (CE), Mean Squared Error (MSE), CORN framework \citep{shi2023deep}, cross-entropy with an ordinal loss term (CO2) \citep{albuquerque2021ordinal}, and ordinal entropy loss (HO2) \citep{albuquerque2021ordinal}. We used Cross Entropy (CE) as our baseline and experimentally evaluated the CDW-CE against other loss functions. %%NEW
The flowchart of the experimental setup is provided in the Figure \ref{fig:flowchart}.
%%NEW

CDW-CE evaluates the performance of a given prediction according to the ordinal nature and penalizes predictions further from the ground truth. The degree of penalization is adjustable through the $\alpha$ hyperparameter, which is a power term above the class distance coefficient. 

\begin{figure}[htbp]
  \centering
  \includegraphics[width=0.9\textwidth]{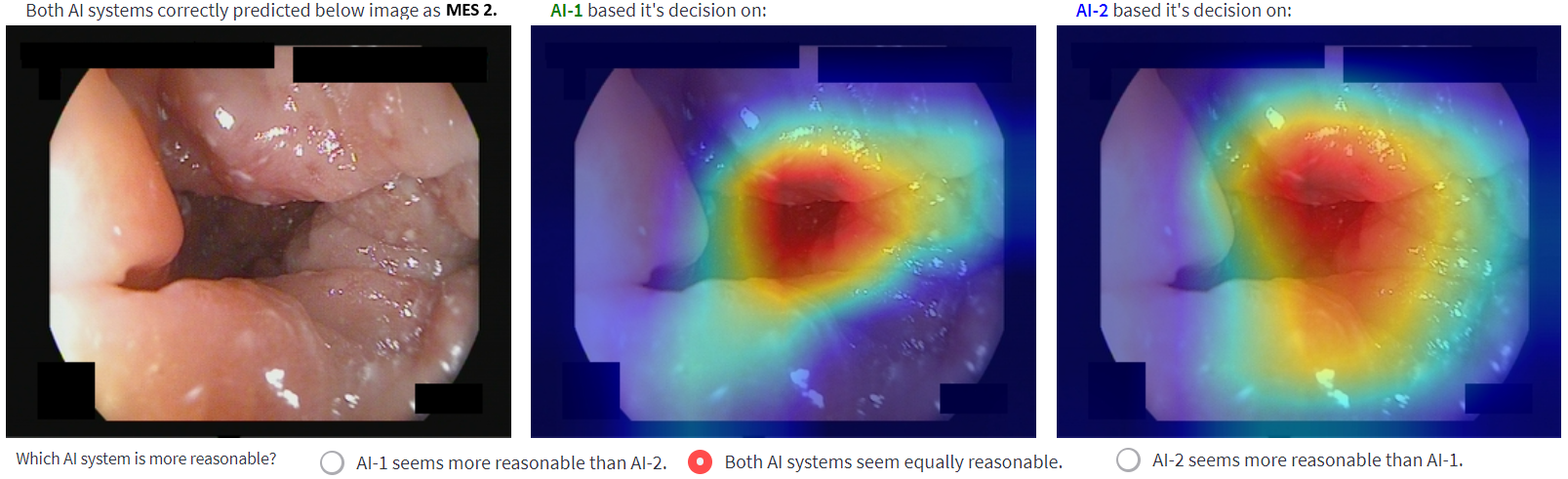}
  
  \caption{A sample CAM output of two different models, trained with and without CDW-CE, presented to medical experts for their feedback.}
  \label{fig:CAM_experiment_screenshot_4}
\end{figure}

The explainability of deep learning models is crucial because they are often deemed black-box systems, given that their features are autonomously learned through an iterative training process. CAMs can be used to visualize and explain where the attention of deep learning models is on the images to overcome this problem. Given a sample input image, CAM demonstrates a deep network's neural activation map in 2-dimensional. We have obtained CAMs of CDW-CE and CE methods and raw endoscopic images and shared them with the medical experts to get their opinion on which model's CAM output is more aligned with symptoms. Figure \ref{fig:CAM_experiment_screenshot_4} provides a sample of CAM visuals.

The ability of a model to extract features correlates with how successfully it has been trained to understand and represent the underlying structure of the data. While feature extraction capability can be illustrated using t-SNE and UMAP plots, these plots do not allow direct numerical comparisons of embedding quality. To adress this, we used the Silhouette Score to quantitatively compare the feature extraction performance of the loss functions. For $N$ samples, the average Silhouette Score is calculated as shown in Eq. \ref{eqn:silhouette}. where $a_i$ is the mean distance from the $i^{th}$ sample to other samples in the same cluster, and $b_i$ is the mean distance from the $i^{th}$ sample to samples in the nearest cluster.

\begin{equation} 
\label{eqn:silhouette}
    S=\frac{1}{N} \sum_{i=1}^{N} \frac{b_i - a_i}{\max(a_i, b_i)}
\end{equation}

\subsection{Dataset}

We conducted our experiments on the LIMUC \citep{LIMUC} dataset, which comprises endoscopic images with varying severity of Ulcerative Colitis (UC), ranked in increasing order from zero to three. UC is one of the two types of inflammatory bowel disease (IBD). Domain experts, i.e., medical doctors, classify whether the disease is present and rate its severity. The corresponding classes of healthy, mild, moderate, and severe disease conditions align with the ordinal levels of Mayo Endoscopic Scoring (MES) of zero to three. While there are several scoring systems for assessing UC from endoscopic images, the Mayo Endoscopic Score (MES) is the most widely used one \citep{satsangi2006montreal}. MES assigns an ordinal label ranging from 0 (healthy instance) to 3 (severe symptoms) for symptom severity assessment of UC. Medical experts also employ a remission state scoring system, consisting of two classes, to distinguish between the presence (MES 2 and 3) and absence (MES 0 and 1) of UC.  

LIMUC dataset is the largest publicly available UC dataset, comprising 11276 images from 1043 colonoscopy procedures on 564 patients. MES classes from 0 to 3 have 6105, 3052, 1254, and 865 images, respectively. The imbalance of the dataset is evident from the percentages corresponding to 54.1\%, 27.1\%, 11.1\%, and 7.7\%, respectively, for each class.
\section{Results} 

We performed experiments for Ulcerative Colitis classification using the dataset detailed in the previous section. The performance results are complemented by an explainability analysis utilizing t-SNE and UMAP plots. Additionally, we gathered feedback from domain experts to evaluate whether our loss function contributes to the model's Class Activation Map (CAM) in better aligning with professional opinion.

\subsection{Assessment of Ulcerative Colitis Classification Performance}

In this study, we have reported Quadratic Weighted Kappa (QWK), F1 Score, accuracy and Mean Absolute Error (MAE) as well as the receiver operating characteristics (ROC) curves with area under the curve (AUC) for true positive and false positive rates. Since precision and recall metrics are used to calculate the F1 score, they were deemed redundant and were not reported separately.

We have employed three deep learning models and six different loss functions, reporting results against various loss functions. The performance results are reported in Table \ref{tab:experiment_results_mayo_scores} and \ref{tab:experiment_results_remission} for multi-class (MES) classification and binary score (remission) classification tasks, respectively.

\begin{table}[ht!]
\centering
\caption{Experiment results for all MES.}
\label{tab:experiment_results_mayo_scores}

\begin{tabular}{lllccc}
\hline
\textbf{}                 & \textbf{Loss} & \textbf{ResNet18}       & \textbf{Inception-v3}   & \textbf{MobileNet-v3-L} \\ \hline
\multirow{6}{*}{QWK}      & CE                              & 0.8296 ± 0.014          & 0.8360 ± 0.011          & 0.8302 ± 0.015              \\
                          & MSE                                        & 0.8540 ± 0.007          & 0.8517 ± 0.007          & 0.8467 ± 0.005              \\
                          & CORN                                       & 0.8366 ± 0.007          & 0.8431 ± 0.009          & 0.8412 ± 0.010              \\
                          & CO2                                        & 0.8394 ± 0.009          & 0.8482 ± 0.009          & 0.8354 ± 0.009              \\
                          & HO2                                        & 0.8446 ± 0.007          & 0.8458 ± 0.010          & 0.8378 ± 0.007              \\
                          & CDW-CE                                     & \textbf{0.8568 ± 0.010} & \textbf{0.8678 ± 0.006} & \textbf{0.8588 ± 0.006}     \\ \hline
\multirow{6}{*}{F1}       & CE                              & 0.6720 ± 0.026          & 0.6829 ± 0.023          & 0.6668 ± 0.028              \\
                          & MSE                                        & 0.6925 ± 0.015          & 0.6881 ± 0.013          & 0.6946 ± 0.011              \\
                          & CORN                                       & 0.6809 ± 0.014          & 0.6832 ± 0.013          & 0.6847 ± 0.020              \\
                          & CO2                                        & 0.6782 ± 0.014          & 0.6846 ± 0.016          & 0.6793 ± 0.012              \\
                          & HO2                                        & 0.6856 ± 0.016          & 0.6901 ± 0.008          & 0.6741 ± 0.030              \\
                          & CDW-CE                                     & \textbf{0.7055 ± 0.021} & \textbf{0.7261 ± 0.015} & \textbf{0.7254 ± 0.010}     \\ \hline
\multirow{6}{*}{Accuracy} & CE                              & 0.7566 ± 0.015          & 0.7600 ± 0.012          & 0.7564 ± 0.011              \\
                          & MSE                                        & 0.7702 ± 0.009          & 0.7690 ± 0.008          & 0.7677 ± 0.009              \\
                          & CORN                                       & 0.7591 ± 0.009          & 0.7600 ± 0.008          & 0.7613 ± 0.012              \\
                          & CO2                                        & 0.7601 ± 0.008          & 0.7654 ± 0.008          & 0.7572 ± 0.009              \\
                          & HO2                                        & 0.7625 ± 0.011          & 0.766 ± 0.010           & 0.7583 ± 0.005              \\
                          & CDW-CE                                     & \textbf{0.7740 ± 0.011} & \textbf{0.7880 ± 0.011} & \textbf{0.7759 ± 0.010}     \\ \hline
\multirow{6}{*}{MAE}      & CE                              & 0.2581 ± 0.018          & 0.2526 ± 0.013          & 0.2563 ± 0.012              \\
                          & MSE                                        & 0.2346 ± 0.009          & 0.2359 ± 0.009          & 0.2383 ± 0.009              \\
                          & CORN                                       & 0.2517 ± 0.012          & 0.2497 ± 0.010          & 0.2480 ± 0.012              \\
                          & CO2                                        & 0.2497 ± 0.011          & 0.2404 ± 0.008          & 0.2524 ± 0.010              \\
                          & HO2                                        & 0.2460 ± 0.011          & 0.2424 ± 0.011          & 0.2487 ± 0.005              \\
                          & CDW-CE                                     & \textbf{0.2300 ± 0.011} & \textbf{0.2147 ± 0.010} & \textbf{0.2272 ± 0.011}     \\ \hline
\end{tabular}
\end{table}

The CDW-CE loss function outperforms all other functions for MES classification by achieving top scores in QWK, F1-Score, accuracy, and MAE metrics. On the other hand, the baseline CE loss function is the least performing among all models. MSE, CORN, CO2, and HO2 consistently outperform CE but are inferior to CDW-CE across various CNN models, with varying performance rankings among themselves.

Classifying disease severity into remission and non-remission holds clinical significance. Consequently, we mapped the results in Table \ref{tab:experiment_results_remission} into remission classification without training a new model. CDW-CE loss function demonstrates the best performance by achieving the highest Kappa, F1-Score, and accuracy metrics. Contrary to multi-class classification, the CE loss function often performs worst among other loss functions while achieving better scores than CORN and CO2 methods for binary classification. Similar to multi-class classification, MSE exhibits second best performance.

\begin{table}[ht!]

\caption{Experiment results for remission classification.}

\label{tab:experiment_results_remission}
\centering
\begin{tabular}{llccc}
\hline
\textbf{}                 & \textbf{Loss} & \textbf{ResNet18}       & \textbf{Inception-v3}   & \textbf{MobileNet-v3-L} \\ \hline
\multirow{6}{*}{Kappa}   & CE          & 0.8077 ± 0.023          & 0.8074 ± 0.021          & 0.8122 ± 0.018              \\
                          & MSE                    & 0.8406 ± 0.013          & 0.8404 ± 0.017          & 0.8339 ± 0.012              \\
                          & CORN                   & 0.8191 ± 0.021          & 0.8077 ± 0.022          & 0.8203 ± 0.016              \\
                          & CO2                    & 0.8185 ± 0.020          & 0.8243 ± 0.011          & 0.8067 ± 0.020              \\
                          & HO2                    & 0.8318 ± 0.015          & 0.8251 ± 0.015          & 0.8283 ± 0.018              \\
                          & CDW-CE                 & \textbf{0.8521 ± 0.016} & \textbf{0.8598 ± 0.012} & \textbf{0.8592 ± 0.012}     \\ \hline
\multirow{6}{*}{F1}       & CE          & 0.8419 ± 0.018          & 0.8420 ± 0.017          & 0.8451 ± 0.016              \\
                          & MSE                    & 0.8691 ± 0.011          & 0.8686 ± 0.014          & 0.8634 ± 0.010              \\
                          & CORN                   & 0.8511 ± 0.016          & 0.8425 ± 0.018          & 0.8523 ± 0.013              \\
                          & CO2                    & 0.8513 ± 0.015          & 0.8561 ± 0.009          & 0.8404 ± 0.017              \\
                          & HO2                    & 0.8618 ± 0.012          & 0.8565 ± 0.011          & 0.8583 ± 0.015              \\
                          & CDW-CE                 & \textbf{0.8785 ± 0.013} & \textbf{0.8847 ± 0.010} & \textbf{0.8842 ± 0.010}     \\ \hline
\multirow{6}{*}{Accuracy} & CE          & 0.9436 ± 0.009          & 0.9432 ± 0.007          & 0.9456 ± 0.005              \\
                          & MSE                    & 0.9531 ± 0.004          & 0.9536 ± 0.006          & 0.9514 ± 0.005              \\
                          & CORN                   & 0.9473 ± 0.007          & 0.9429 ± 0.008          & 0.9473 ± 0.006              \\
                          & CO2                    & 0.9461 ± 0.008          & 0.9479 ± 0.004          & 0.9444 ± 0.006              \\
                          & HO2                    & 0.9507 ± 0.005          & 0.9485 ± 0.005          & 0.9504 ± 0.005              \\
                          & CDW-CE                 & \textbf{0.9566 ± 0.005} & \textbf{0.9590 ± 0.003} & \textbf{0.9588 ± 0.005}     \\ \hline
\end{tabular}

\end{table}

\begin{figure}[!ht]
\begin{subfigure}[b]{0.49\textwidth}
\centering
  \includegraphics[width=1\textwidth]{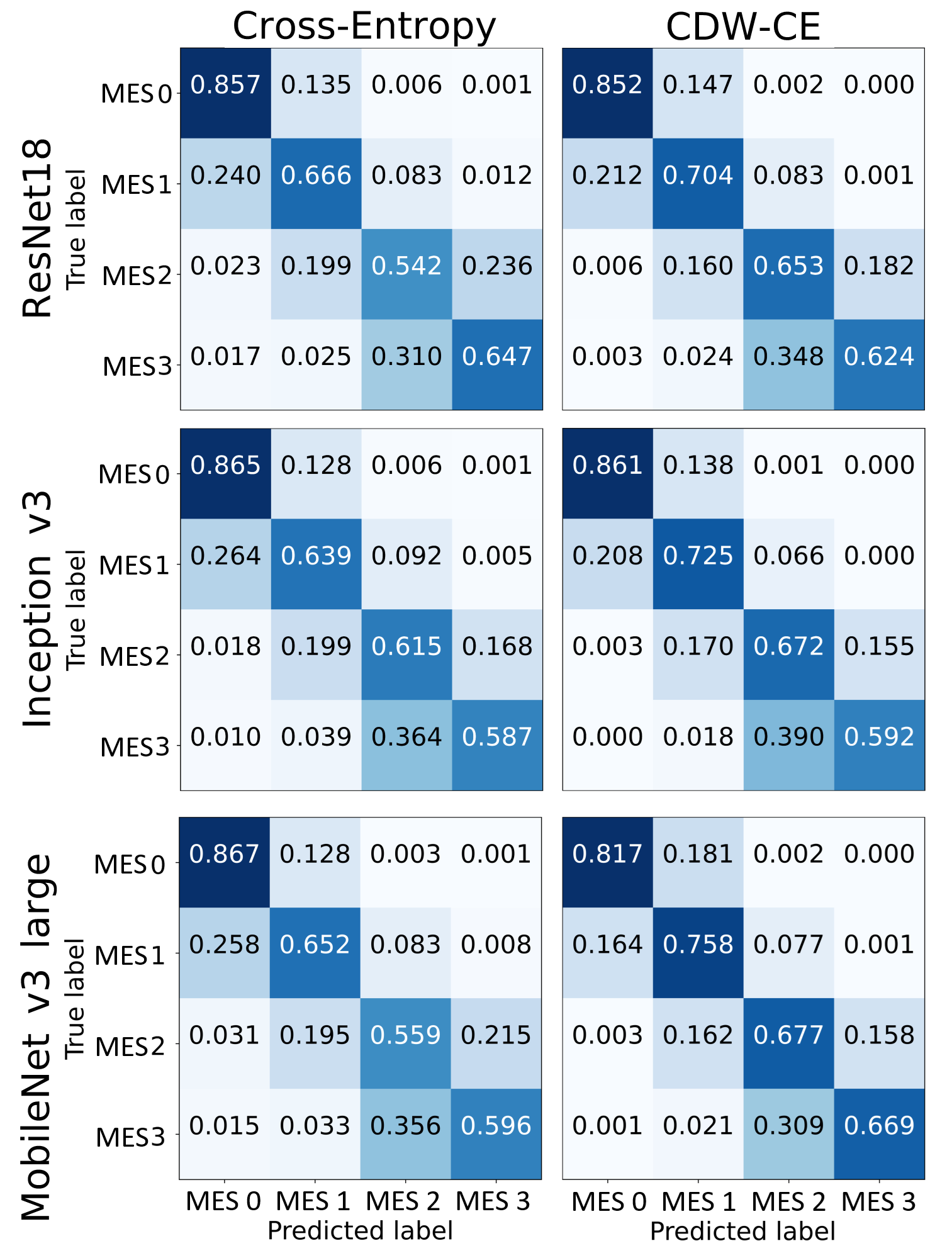}
  \caption{Mean confusion matrix for full MES classification.}
  \label{fig:confusion_matrix_4_classes}
\end{subfigure}
  \hfill
  \begin{subfigure}[b]{0.49\textwidth}
  \centering
    \includegraphics[width=0.82\textwidth]{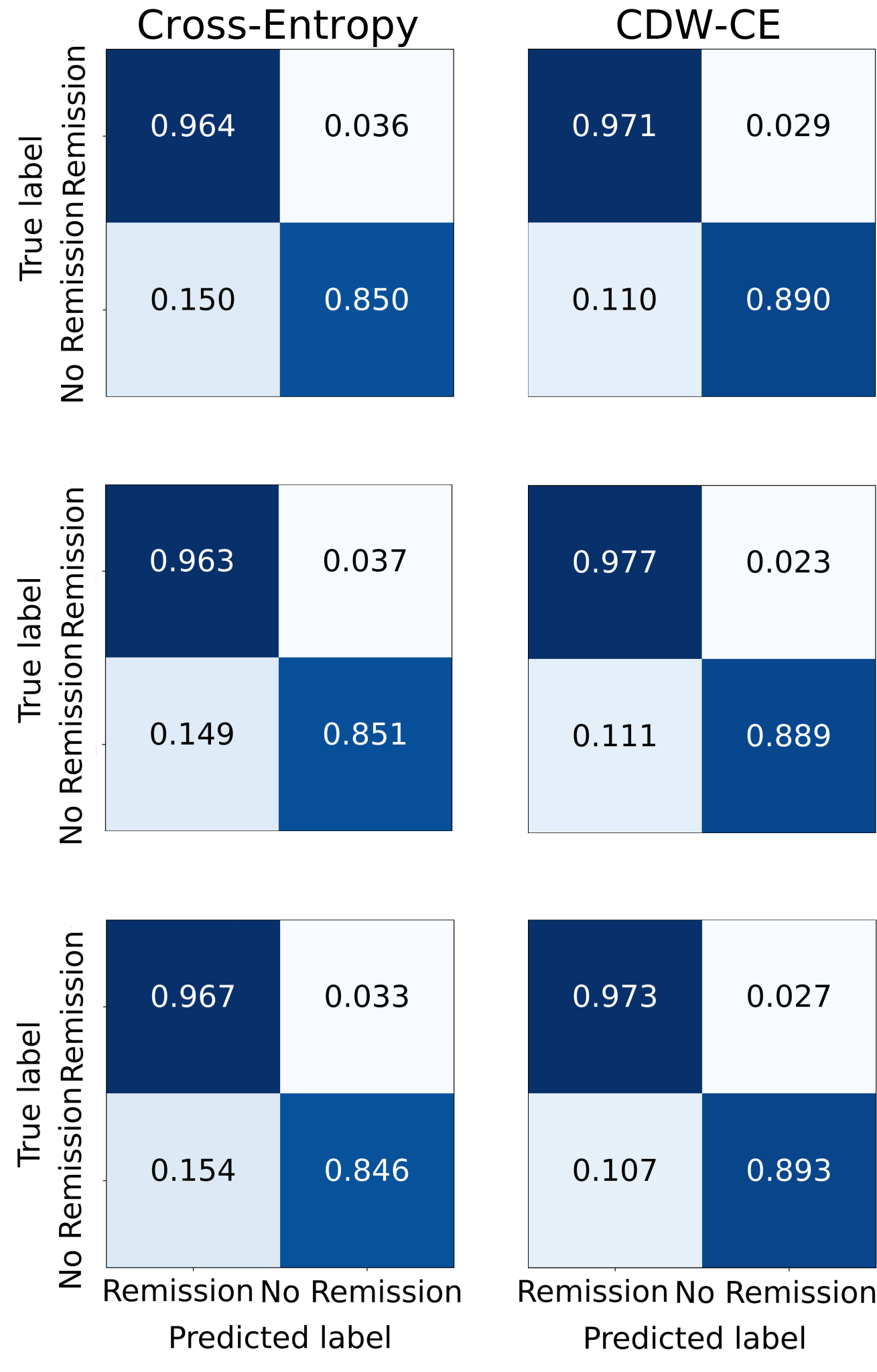}
  \caption{Mean confusion matrix for remission.}
  \label{fig:confusion_matrix_remission}
  \end{subfigure}
  \caption{Mean confusion matrix of each CNN model trained with CE and CDW-CE for full MES classification.}
\end{figure}

The confusion matrices for multi-class and binary classification tasks are shown in Figures \ref{fig:confusion_matrix_4_classes} and \ref{fig:confusion_matrix_remission}, respectively. Figure \ref{fig:confusion_matrix_4_classes} demonstrates that the CDW-CE loss function decreases mispredictions with a distance larger than one. With decreased mispredictions for two and three-class distances, the loss function manipulates the model learning as expected. Since CDW-CE clusters mispredictions around 1-class distance, naturally, this is also reflected in the better classification as shown in Figure \ref{fig:confusion_matrix_remission}. \\

\begin{figure}[!ht]
    \centering
    \begin{subfigure}[b]{0.48\textwidth}
        \centering
        \includegraphics[width=\textwidth,trim={0.5cm 0 1.4cm 0.5cm},clip]{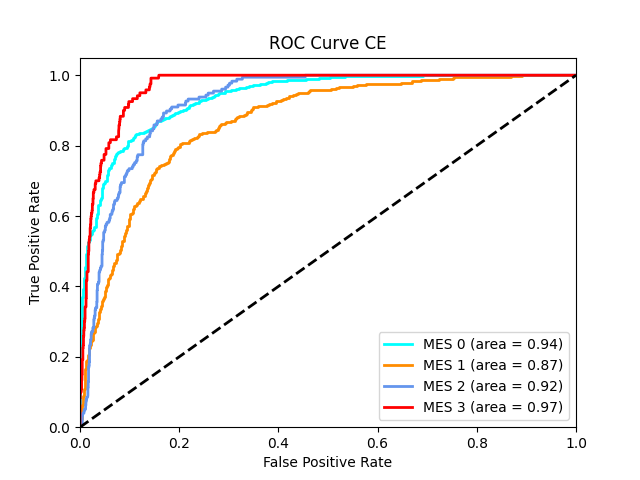}
        %\caption{ROC curve for loss function CE}
        %\label{fig:roc_ce}
    \end{subfigure}
    %\hfill % Important for spacing
    \begin{subfigure}[b]{0.48\textwidth}
        \centering
        \includegraphics[width=\textwidth,trim={0.5cm 0 1.4cm 0.5cm},clip]{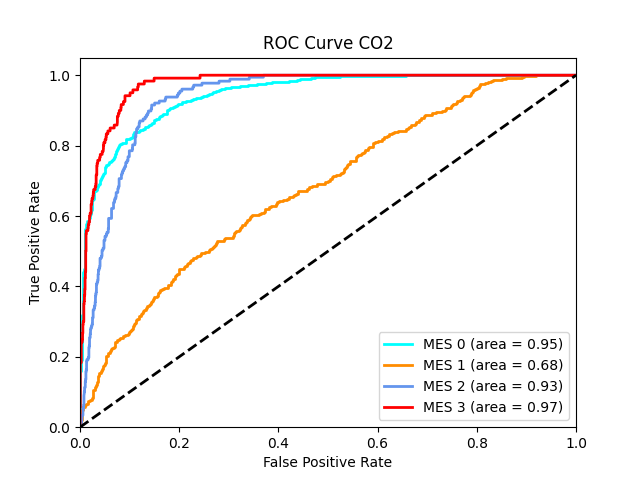}
        %\caption{ROC curve for loss function CO2}
        %\label{fig:roc_co2}
    \end{subfigure}
    \\%[\smallskip] % Add a small vertical space between rows
    \begin{subfigure}[b]{0.48\textwidth}
        \centering
        \includegraphics[width=\textwidth,trim={0.5cm 0 1.4cm 0.5cm},clip]{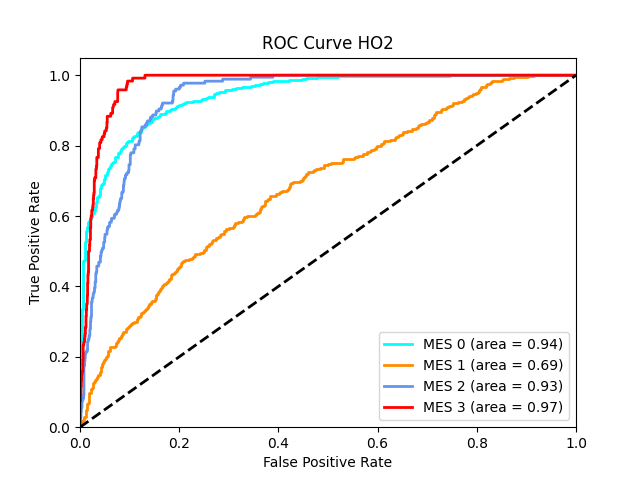}
        %\caption{ROC curve for loss function HO2}
        %\label{fig:roc_ho2}
    \end{subfigure}
    %\hfill
    \begin{subfigure}[b]{0.48\textwidth}
        \centering
        \includegraphics[width=\textwidth,trim={0.5cm 0 1.4cm 0.5cm},clip]{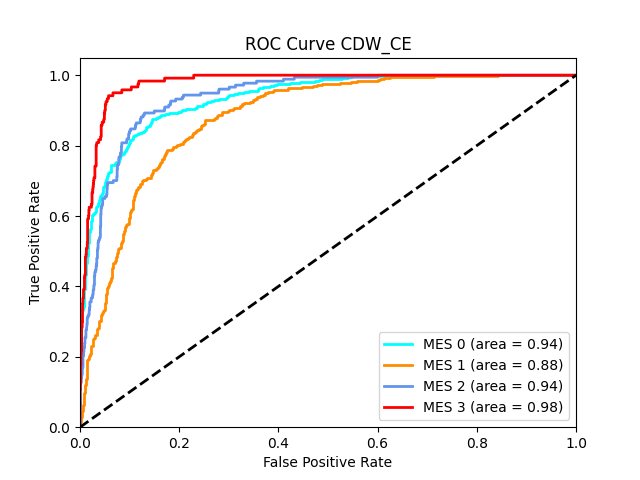}
        %\caption{ROC curve for loss function CDW-CE}
        %\label{fig:roc_cdwce}
    \end{subfigure}
    \caption{ROC curves obtained with the same deep learning architecture trained with different loss functions.}
    \label{fig:rocs}
\end{figure}

Receiver operating characteristic (ROC) curves measure the trade-off between false positive and true positive rates, with the area under the curve (AUC) indicating a model’s ability to discriminate between classes. Figure {\ref{fig:rocs}} presents the ROC curves and corresponding AUC scores for each MES class,based on the ResNet18 architecture trained with various loss functions.

The ROC curves for various loss functions are presented in Figure {\ref{fig:rocs}}. The results show that CO2 and HO2 exhibit significantly worse performance for the MES 1 class, while achieving similar or better performance for other classes compared to the CE loss function. The model trained with CDW-CE outperforms all others, including ordinal loss functions, by achieving higher AUC scores across all classes. Analyzing the ROC curves for different loss functions reveals that CDW-CE not only improves classification performance in terms of QWK and F1 scores but also improves the balance between false positive and true positive rates.

\section{Discussion}
\subsection{Explainability Analysis using t-SNE plots}

In this section, we use t-SNE \citep{van2008visualizing} and UMAP \citep{McInnes2018} plots to visually assess the feature extraction capability of our method by examining high-dimensional data. UMAP and t-SNE effectively reveals clusters and latent patterns in the data. To complement this, we analyze silhouette scores, which quantify clustering quality by measuring intra-class cohesion and inter-class separation. A higher silhouette score indicates better clustering. While t-SNE and UMAP plots are helpful for visual insights, silhouette scores provide an objective comparison of different approaches.
\begin{figure}[htbp]
\centering
    \begin{subfigure}[b]{0.36\textwidth}
        \includegraphics[width=1.1\textwidth,trim={1cm 0.2cm 0.2cm 0.5cm},clip]{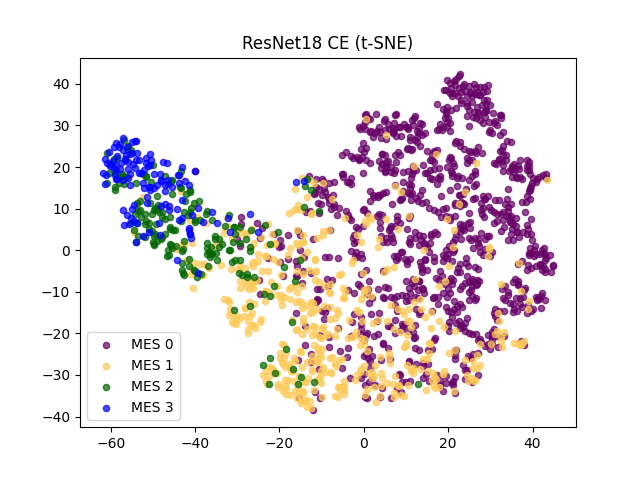}
        %\caption{CE t-SNE mean Silhouette Score: 0.156}
        %\label{fig:tsne_ce}
    \end{subfigure}
    \begin{subfigure}[b]{0.36\textwidth}
        \centering
        \includegraphics[width=1.1\textwidth,trim={1cm 0.2cm 0.2cm 0.5cm},clip]{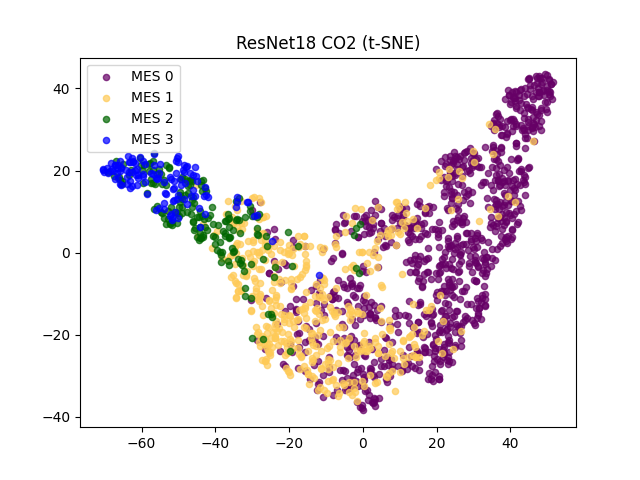}
        %\caption{CO2 t-SNE mean Silhouette Score: 0.138}
        %\label{fig:tsne_co2}
    \end{subfigure}
    \\ 
    \begin{subfigure}[b]{0.36\textwidth}
        \includegraphics[width=1.1\textwidth,trim={1cm 0.2cm 0.2cm 0.5cm},clip]{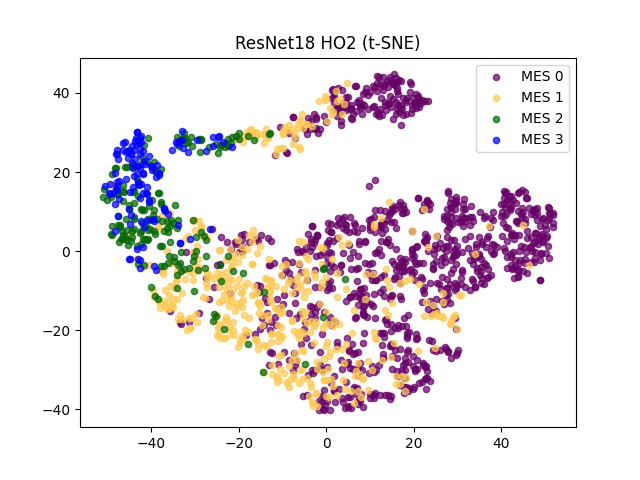}
        %\caption{HO2 t-SNE mean Silhouette Score: 0.092}
        %\label{fig:tsne_ho2}
    \end{subfigure}
    \begin{subfigure}[b]{0.36\textwidth}
        \centering
        \includegraphics[width=1.1\textwidth,trim={1cm 0.2cm 0.2cm 0.5cm},clip]{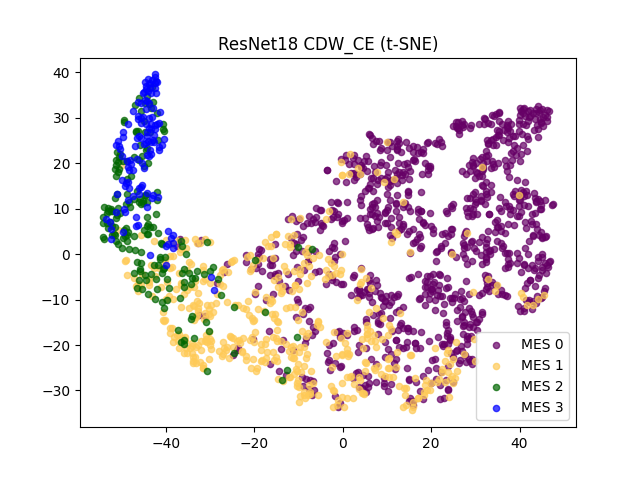}
        %\caption{CDW-CE t-SNE mean Silhouette Score: 0.176}
        %\label{fig:tsne_cdwce}
    \end{subfigure}
    \caption{t-SNE plots obtained with the same deep learning architecture trained with different loss functions.}
    \label{fig:tsne}
\end{figure}

\begin{figure}[htbp]
\centering
    \begin{subfigure}[b]{0.36\textwidth}
        \includegraphics[width=1.1\textwidth,trim={1cm 0.2cm 0.2cm 0.5cm},clip]{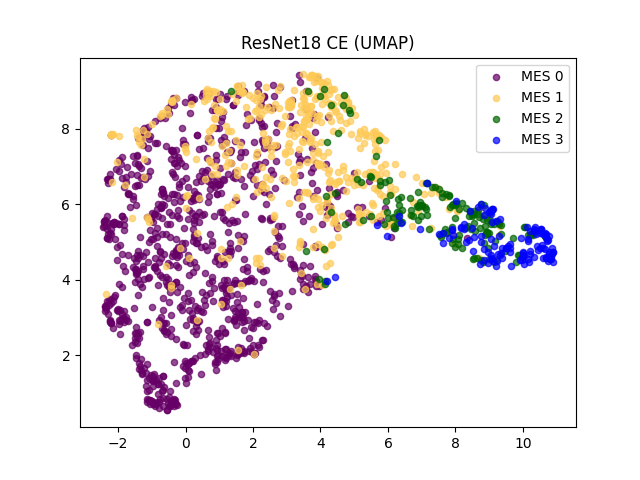}
        %\caption{CE UMAP Silhouette Score: 0.167}
        %\label{fig:umap_ce}
    \end{subfigure}
    \begin{subfigure}[b]{0.36\textwidth}
        \centering
        \includegraphics[width=1.1\textwidth,trim={1cm 0.2cm 0.2cm 0.5cm},clip]{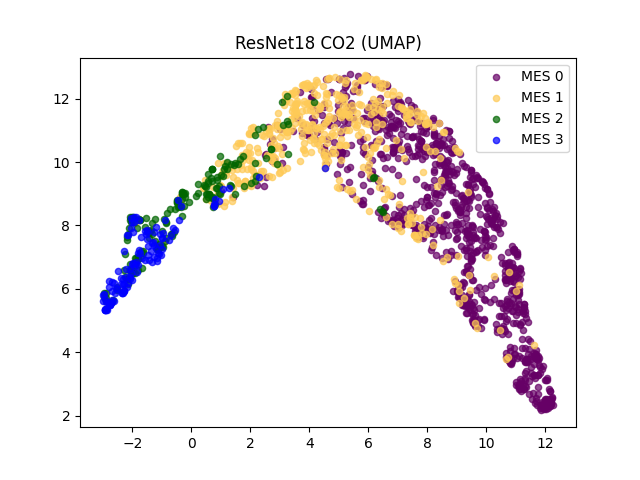}
        %\caption{CO2 UMAP Silhouette Score: 0.163}
        %\label{fig:umap_co2}
    \end{subfigure}
    \\
    \begin{subfigure}[b]{0.36\textwidth}
        \includegraphics[width=1.1\textwidth,trim={1cm 0.2cm 0.2cm 0.5cm},clip]{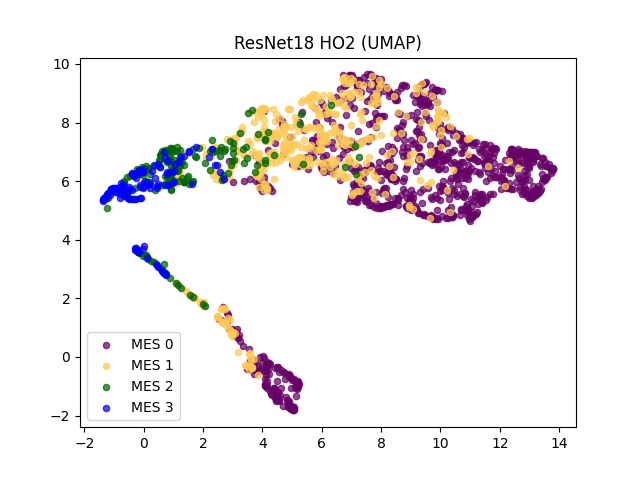}
        %\caption{HO2 UMAP Silhouette Score: 0.089}
        %\label{fig:umap_ho2}
    \end{subfigure}
    \begin{subfigure}[b]{0.36\textwidth}
        \centering
        \includegraphics[width=1.1\textwidth,trim={1cm 0.2cm 0.2cm 0.5cm},clip]{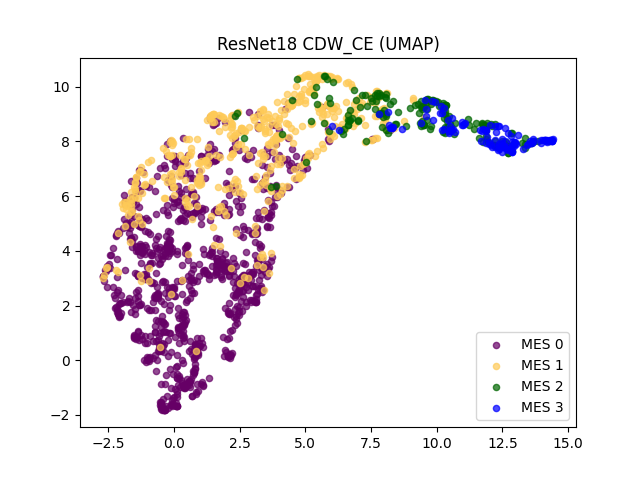}
        %\caption{CDW-CE UMAP Silhouette Score: 0.201}
        %\label{fig:umap_cdwce}
    \end{subfigure}
    \caption{UMAP plots obtained with the same deep learning architecture trained with different loss functions.}
    \label{fig:umap}
\end{figure}

Figures \ref{fig:tsne} and \ref{fig:umap} display the t-SNE and UMAP plots illustrating the embeddings of our models trained on LIMUC using CE and CDW-CE, respectively. The plots reveal a notable separation between MES 0 and MES 3 classes compared to MES 1 and MES 2 classes.

The mean silhouette scores for models trained with various loss functions are presented in Table {\ref{tab:silhouette_scores}}. The results indicate that models trained using CDW-CE achieve higher silhouette scores on both t-SNE and UMAP plots, suggesting better feature extraction capabilities than models trained with other loss functions.

\begin{table}[htbp]
    \centering
    \begin{tabular}{|c|c|c|}\hline
      Loss Function     &  t-SNE                & UMAP              \\\hline
      CE	              &  0.156                & 0.167             \\\hline
      CO2	              &  0.138                & 0.163             \\\hline
      HO2	              &  0.092                & 0.089             \\\hline
      CDW-CE            &  \textbf{0.176}       & \textbf{0.201}    \\\hline
    \end{tabular}
    \caption{Mean silhouette scores of t-SNE and UMAP plots for models trained with different loss functions.}
    \label{tab:silhouette_scores}
\end{table}

\subsection{Explainability Analysis: Class Activation Maps}

\begin{figure}[ht!]
  \centering
  \includegraphics[width=1\textwidth, trim= {0 1cm 0 1cm}]{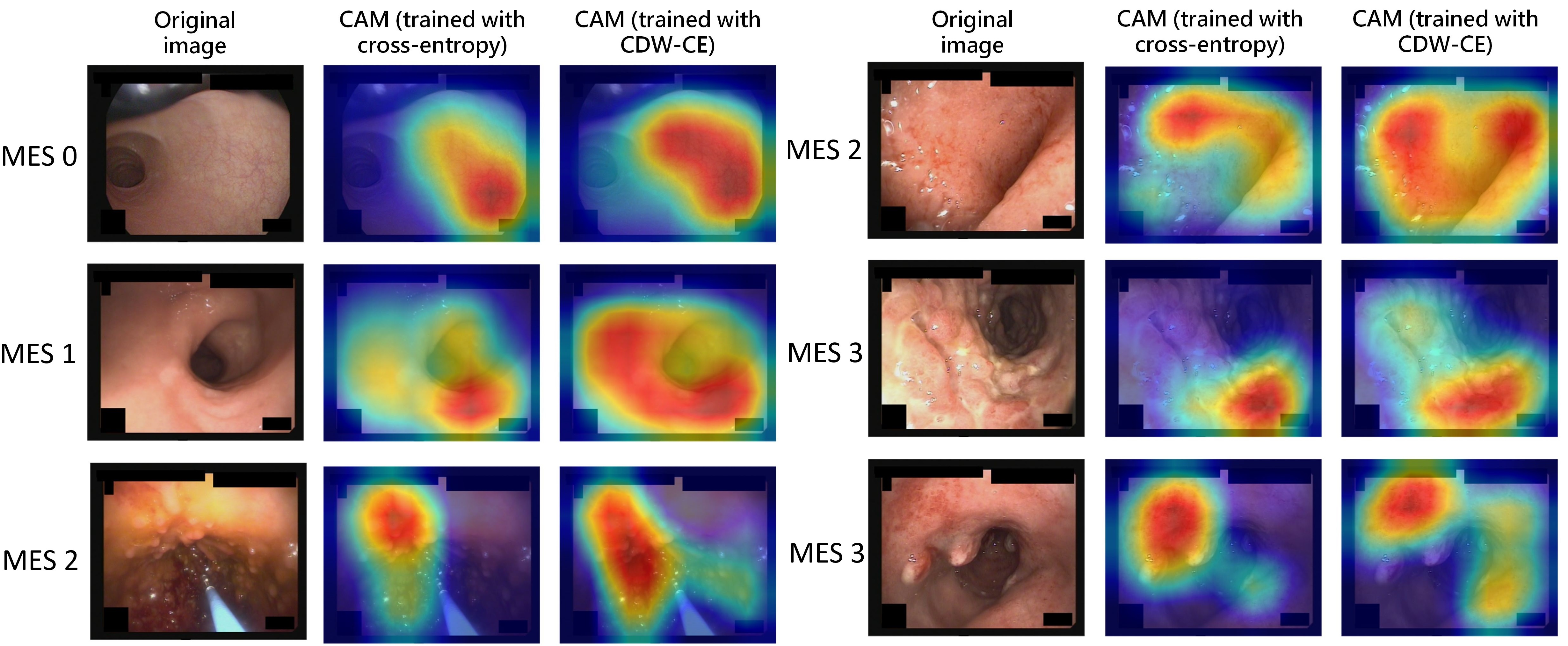}
  \caption[ResNet18 model CAM visualizations.]{Original images and CAM visualizations of the ResNet18 model trained with CE and CDW-CE. The model trained with CDW-CE highlights broader and more relevant areas associated with the disease.}
  \label{fig:sample_cams}
\end{figure}

Class Activation Maps (CAM) enable the analysis of where the model focuses its attention in a given input image. In this study, we obtained CAM for two models—one trained using our CDW-CE method and the other with the CE loss function. Some sample images and the corresponding CAM visuals for both models are provided in Figure \ref{fig:sample_cams}. 

\begin{figure}[ht!]
  \centering
  \includegraphics[width=350pt, keepaspectratio]{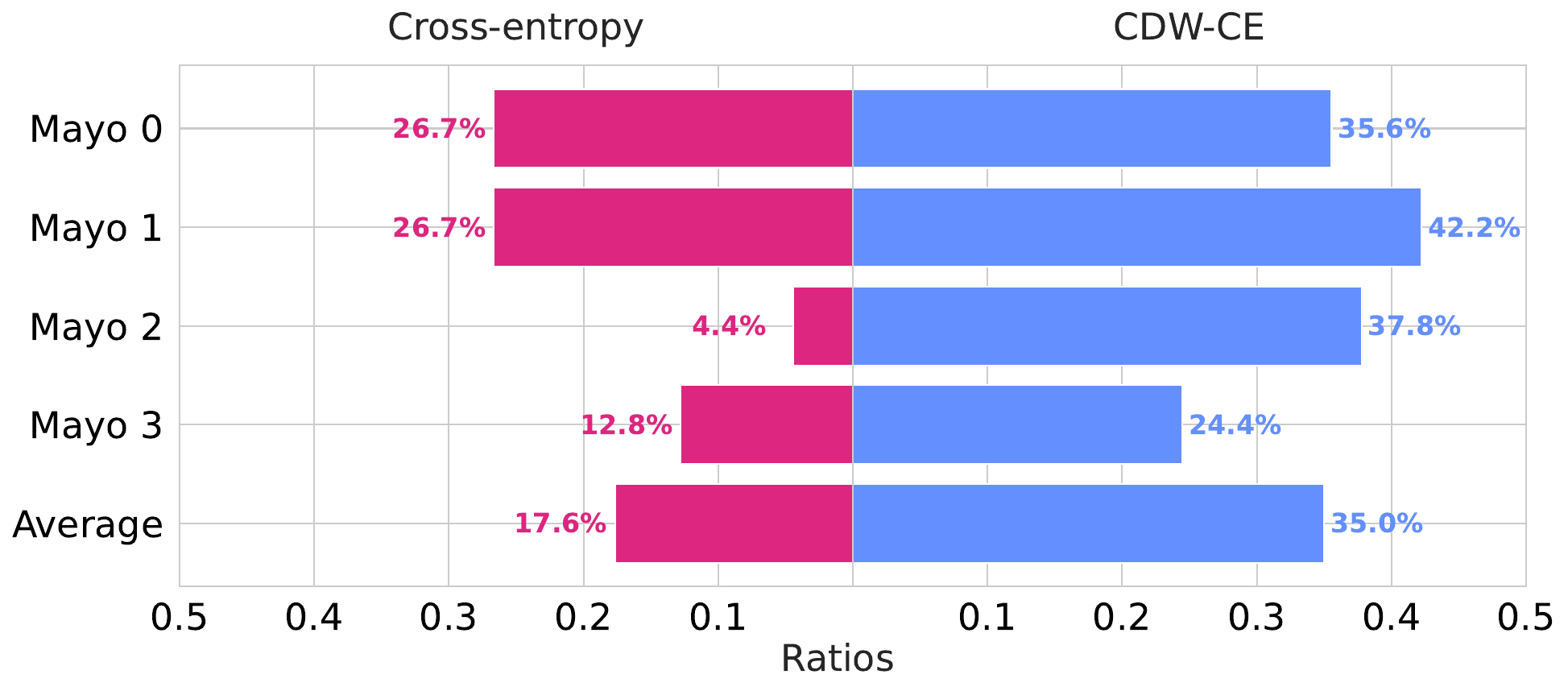}
  \caption[The assessment results of CAM visualizations for models trained with CE and CDW-CE.]{Expert evaluations of CAM visualizations for models trained with CE and CDW-CE. The percentage values representing the instances where experts found both visualizations equally reasonable were 37.7\%, 31.1\%, 57.8\%, 62.8\%, and 47.4\%, respectively.}
  \label{fig:cam_comparsion_result}
\end{figure}

CAM visualizations can be employed to obtain important feedback from domain experts through a simple questionnaire, asking them to identify which model's attention aligns more accurately with the underlying symptoms. For this task, we presented the experts with 240 images, comprising 60 images from each category. Both models correctly identified these images. The experts were only shown the original image and two CAM visualizations superimposed on the original images. The CAM images generated by the models for every new image were arbitrarily labeled as AI-1 (Artificial Intelligence-1) and AI-2. Without being aware of which model produced which CAM visualization, the clinicians were requested to choose which CAM visualization was more aligned with their decision-making. If there were no significant differences, they had the option to indicate that CAM images were equally reasonable in terms of decision-making. According to their feedback, as can be observed from Figure \ref{fig:cam_comparsion_result}, our method enables the CNN models to better align with the expectations of the domain experts on where the model shall attend.

\subsection{Analysis of the Hyperparameters}

The power term $\alpha$ in the loss function controls the strength of the penalization, with higher values leading to stronger penalization for mispredictions that deviate further from the true class. We have integrated a margin term to the CDW-CE loss function to penalize inconsistent classifications further and make the classifier more robust regarding better classification boundaries and a more considerable distance between clusters of classes. Adding these terms aims to enhance intra-class compactness, leading to more tightly grouped in-class distributions. Similarly, the inter-class distance, or the separation between different class clusters, is intended to be improved. While this results in better classification performance and robustness, it also requires more extensive hyperparameter tuning.

\begin{figure}[ht!]
  \centering
  \includegraphics[width=\textwidth,trim={0 0 0 1cm}]{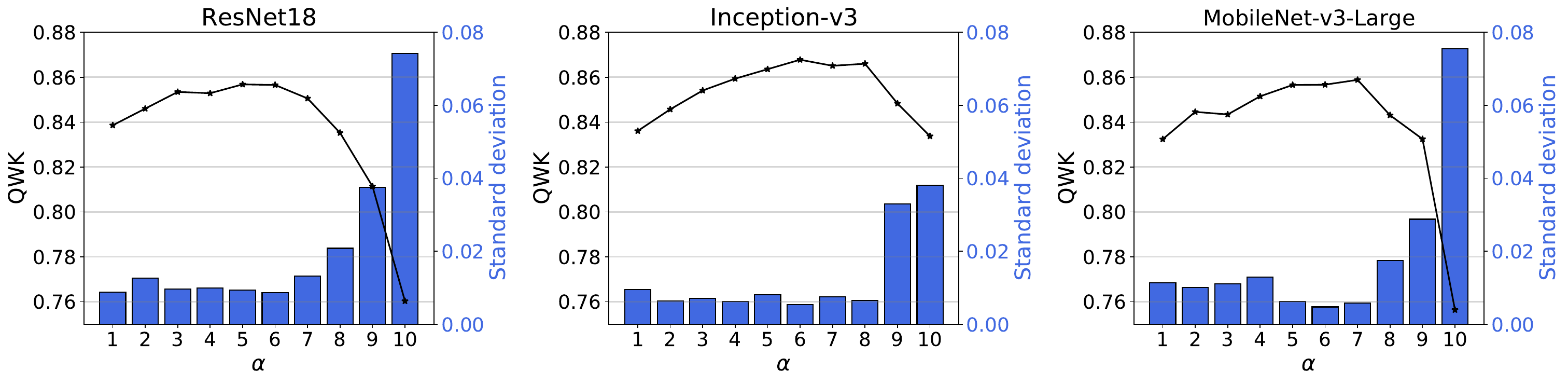}
  \caption{Effects of $\alpha$ for three different models on the mean and standard deviation of QWK scores.}
  \label{fig:model_power_analysis}
\end{figure}

We analyzed the impact of $\alpha$ by varying it across integer values within the range of 1 to 10. For each model (ResNet18, Inception-v3, and MobileNet-v3-large), we found the best-performing power term values as 5, 6, and 7, respectively. These experiments are summarized in the Figure \ref{fig:model_power_analysis}.

\begin{figure}[ht!]
  \centering
  \includegraphics[width=0.6\textwidth,trim={0 0 0 1.3cm}, clip]{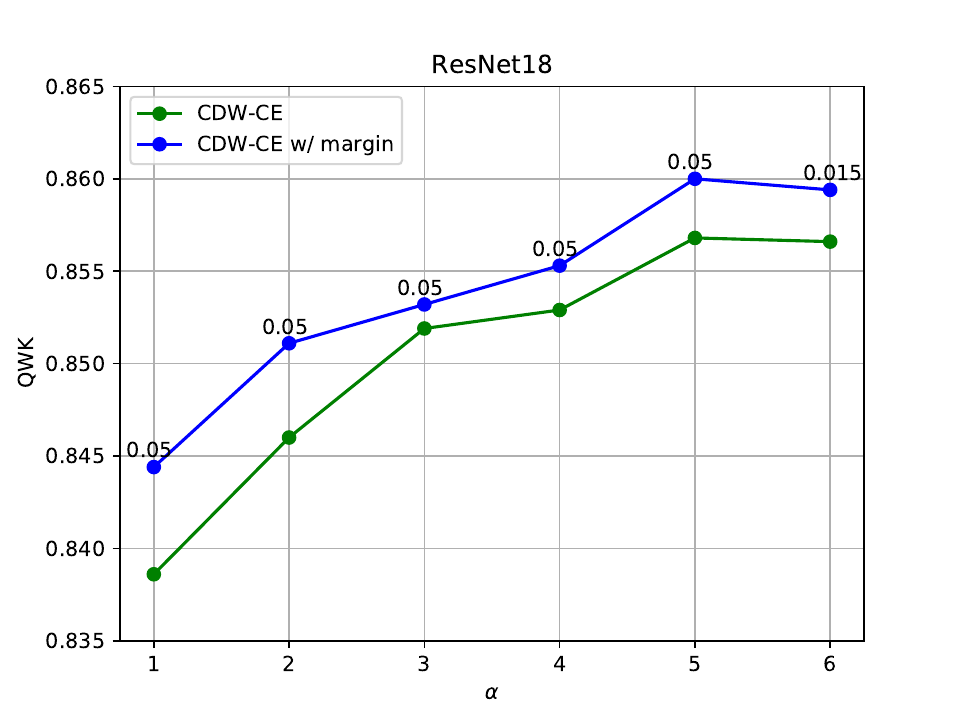}
  \caption[CDW-CE vs. CDW-CE with margin for varying $\alpha$ values.]{CDW-CE vs. CDW-CE with margin for varying $\alpha$ values. Numbers above the blue markers show the margin value.}
  \label{fig:cdw_ce_with_margin_result}
\end{figure}

The addition of the margin term improves the CDW-CE loss function in terms of classification performance as demonstrated by a ResNet18 architecture trained with and without margin term CDW-CE loss functions with different $\alpha$ values in Figure \ref{fig:cdw_ce_with_margin_result}. Here, we have approached the margin term as a new hyperparameter and tuned it according to the QWK performance. With this tuning, we have achieved improved results for different margin values corresponding to different power terms ($\alpha$). The $\alpha = 1$ is the basic CDW-CE loss function with no class distance weights, and the inclusion of the margin term improves the results in this baseline form.

Adding a margin term effectively improves all models used in this work in terms of QWK score as shown in Table \ref{tab:cdw_ce_m_comparison}.

\begin{table}[ht!]
\centering
\caption[QWK scores for CDW-CE vs. CDW-CE with margin]{QWK scores for CDW-CE vs. CDW-CE with margin ($m$ refers to additive margin value).}
\label{tab:cdw_ce_m_comparison}
\begin{tabular}{lccc}
\multicolumn{1}{c}{\textbf{CDW-CE}} & \textbf{ResNet18}   & \textbf{Inception-v3}    & \textbf{MobileNet-v3-L} \\ \hline
no margin                                     & 0.8568              & 0.8678                         & 0.8588               \\
with margin                           & 0.8600 ($m$=0.05)   & 0.8719 ($m$=0.025)             & 0.8617 ($m$=0.0025)     
\end{tabular}
\end{table}

\section{Conclusions}  

In this work, we have proposed a new loss function, Class Distance Weighted Cross-Entropy (CDW-CE), designed specifically for ordinal classification tasks. Our experiments on the LIMUC dataset demonstrated that ordinal loss functions perform better than the widely used non-ordinal loss function CE and other ordinal loss functions. Moreover, the proposed CDW-CE method outperformed the other methods across various performance metrics and architectures. With ROC curves and corresponding AUC scores, CDW-CE outperformed other loss functions. Additionally, an analysis of remission scores indicated that CDW-CE surpasses other ordinal loss functions. 

To demonstrate the encoding capability of models trained with our approach, we generated t-SNE and UMAP plots. We calculated corresponding silhouette scores for CE, HO2, CO2 and CDW-CE loss functions, representing ordinal and categorical scenarios. The Silhouette Score comparison revealed that CDW-CE, with higher scores, exhibits superior intra-class grouping and better inter-class separation. Furthermore, feedback from domain experts based on CAM visualizations indicated that our model outperformed the baseline CE loss function, suggesting improved explainability.

\section{Acknowledgements}

This work has been supported by Middle East Technical University Scientific Research Projects Coordination Unit under grant number ADEP-704-2024-11486. 

\section{Author contributions}

All authors took part in the conceptualization. Gorkem Polat performed data curation, formal analysis, investigation, validation, visualization and contributed to the methodology, software and revision of the manuscript. Ümit Mert Çağlar contributed to the data curation, software, visualization, preparation of the original draft and editing the manuscript. Alptekin Temizel provided resources, funding acquisition, supervision, project administration and contributed to the methodology, validation and review of the manuscript.

\section{Data availability} 

Labeled Images for Ulcerative Colitis (LIMUC) dataset is a publicly available dataset, accessible at https://zenodo.org/records/5827695.

\section{Conflicts of interest}
The authors declare that they have no known competing financial interests or personal relationships that could have appeared to influence the work reported in this paper.

\bibliographystyle{elsarticle-harv} 
\bibliography{main}

\end{document}